\documentclass[article]{elsarticle}

\usepackage{lineno,hyperref}
\usepackage{lipsum}
\usepackage{amsmath,amssymb,amsfonts}
\usepackage{algorithmic}
\usepackage{graphicx}
\usepackage{textcomp}
\usepackage{float}
\usepackage{wrapfig}
\usepackage{lscape}
\usepackage{rotating}
\usepackage{epstopdf}
\usepackage{multirow}
\modulolinenumbers[1]

\journal{Journal of Medical Systems}

\def\BibTeX{{\rm B\kern-.05em{\sc i\kern-.025em b}\kern-.08em
    T\kern-.1667em\lower.7ex\hbox{E}\kern-.125emX}}

\begin{document}

\begin{frontmatter}

\title{Generative Adversarial Network for Medical Images (MI-GAN)}

\author{Talha Iqbal}
\address{Department of Electrical Engineering, COMSATS University Islamabad, Abbottabad Campus, Pakistan.}
\author{Hazrat Ali}
\address{Department of Electrical Engineering, COMSATS University Islamabad, Abbottabad Campus, Pakistan.}
\cortext[mycorrespondingauthor]{Corresponding author: Hazrat Ali, Department of Electrical Engineering, COMSATS University Islamabad, Abbottabad Campus, Pakistan.}
\ead{hazratali@ciit.net.pk}

\begin{abstract}
Deep learning algorithms produces state-of-the-art results for different machine learning and computer vision tasks. To perform well on a given task, these algorithms require large dataset for training. However, deep learning algorithms lack generalization and suffer from over-fitting whenever trained on small dataset, especially when one is dealing with medical images. For supervised image analysis in medical imaging, having image data along with their corresponding annotated ground-truths is costly as well as time consuming since annotations of the data is done by medical experts manually. In this paper, we propose a new Generative Adversarial Network for Medical Imaging (MI-GAN). The MI-GAN generates synthetic medical images and their segmented masks, which can then be used for the application of supervised analysis of  medical images. Particularly, we present MI-GAN for synthesis of retinal images. The proposed method generates precise segmented images better than the existing techniques. The proposed model achieves a dice coefficient of 0.837 on STARE dataset and 0.832 on DRIVE dataset which is state-of-the-art performance on both the datasets.
\end{abstract}

\begin{keyword}
GAN. \sep medical imaging. \sep style transfer. \sep deep learning. \sep retinal images. 
\end{keyword}

\end{frontmatter}


\section{Introduction}
In recent times, strong interest has emerged in the use of computer-aided medical diagnosis \cite{a1} \cite{a2} \cite{a3} \cite{a4} \cite{a5} \cite{a6} \cite{a7} \cite{a8} \cite{a9} \cite{a10}. Computer aided diagnosis relies on advanced machine learning and computer vision techniques \cite{b1}. Today, majority of the medical professionals use computer-aided medical images for diagnosis purposes. Retinal vessel network analysis gives us information about the status of general system and conditions of the eyes. Ophthalmologists can diagnose early sign of vascular burden due to hypertension and diabetes as well as vision threatening retinal diseases like Retinal Artery Occlusion (RAO) and Retinal Vein Occlusion (RVO) from abnormality in vascular structure \cite{b36}. To aid this kind of analysis, automatic vessels segmentation methods have been extensively studied. Recently, deep learning methods have shown potential to produce promising results with higher accuracy, occasionally better than medical specialist in the field of medical imaging \cite{b3}. Deep learning also improves efficiency of analyzing data due to its computational and automated nature but most of the medical images are often 3 dimensional (e.g. MRI and CT) and it is difficult as well as inefficient to produce manually annotated images. In general, medical images are inadequate, expensive and offer restricted use due to legal issues (patient privacy). Moreover, the datasets of medical images available publicly often lack consistency in size and annotation. This makes them less useful for training of neural networks, which are data-hungry. This directly limits the development of medical diagnosis systems. So, generation of synthetic images along with their segmented images will help in medical image analysis and provide better diagnosis systems.\\
Recent work in the domain of medical imaging has shown possibility of improved performance even on small datasets. This has became possible through provision of some prior knowledge in a deep neural network \cite{b2}. U-net \cite{b3} architecture is popular for segmentation of bio-medical images, which shows how strongly an augmented data can be utilized to cope with low amount of training data available to train deep networks. Data augmentation is easy to implement and gives good results but it is only able to give fixed variations for any given dataset and requires the augmentation to fit in the given dataset. Impressive results are achieved by Gatys et al. \cite{b6} by  application of deep learning algorithm. Similar approach with modifications has been used by \cite{b7}, \cite{b8}, reducing the computational complexity. More traditional approaches for segmentation of filamentary structured images have been reported in \cite{b9} and \cite{b10}.\\
Generative Adversarial Networks (GANs) are useful for many applications like unsupervised representation learning \cite{b4} or image-to-image translation \cite{b5}. Typically, vessel segmentation task is considered as image translation problem where segmented vessel map is produced at output using fundoscopic image as an input to the model. We can have clearer and sharper vessel segmented masks, if we constrain our output to resemble the annotation done by human experts. For image generation, Generative Adversarial Networks (GANs) \cite{b18} provide a different approach. GANs are divided into two networks i.e. Generator and Discriminator. Both are trained to compete with each other like min-max game. Goal of discriminator is to classify the input image as real or synthetic image while generator goal is to generate images that are close to real so that discriminator gets fooled by it. To deal with over-fitting, generator is never shown the training dataset and is only fed with the gradient of discriminator decision. The training process is highly affected by the values of hyper-parameters. The major problem in GANs is to find Nash Equilibrium to stop the training process of generator and discriminator, which can otherwise lead to training instability.\\
Number of GANs like \cite{b11}, \cite{b12}, \cite{b13}, \cite{b14} have been developed. DCGAN \cite{b11} introduced set of constraints which stabilized the training of the model. CGAN \cite{b12} trained the model and generated output conditioned to some auxiliary information. LAPGAN \cite{b13} uses cascade formation of convolutional neural networks within framework of Laplacian pyramid for the generation of the new images. InfoGAN \cite{b14} learns disentangled representations in unsupervised manner. GANs have performed well on small medical image datasets as discussed in \cite{b19}. The authors in \cite{b19} have used GANs for unsupervised adaptation of the multi-model medical images. \\
In this paper, we propose a new approach for generation of \textbf{retinal vessel images} as well as their segmented masks using generative adversarial networks. The closest to our work is that of \cite{b15}. The method proposed in \cite{b15} is limited to generation of fixed output for a given input. On the contrary, our method can produce unlimited number of synthetic images from same input. Moreover, unlike \cite{b15} that uses hundred to millions of training examples, our approach works on only tens of training images. Our method not only extracts sharp and clearer vessels having less false positives as compared to existing methods but also achieve state-of-the-art performance on two publicly available datasets i.e. STARE \footnote{http://cecas.clemson.edu/~ahoover/stare/}and DRIVE \footnote{https://www.isi.uu.nl/Research/Databases/DRIVE/}. Our model, when trained on the generated datasets, gives comparable results with the network trained on real data images. The major contributions of this work are:
\begin{itemize}
\item We propose a GAN which is able to generate realistic looking retinal images from only tens of examples, unlike \cite{b15}, which requires hundreds of training examples.
\item We propose a variant of the style transfer based on particular style representation provided by additional input.
\item Unlike the traditional training of GANs, we propose a new technique. We update generator twice than discriminator to get quicker convergence. Thus, the overall training time is reduced significantly.
\end{itemize}
The rest of the paper is organized as follows:
We explain Generative Adversarial Network and the proposed design of our model in Section II. We have discussed experimental setup and results in Section III. Finally, the paper is concluded in Section IV.
\begin{figure*}[]
  \begin{center}
	\includegraphics[width=\textwidth,height=8cm, keepaspectratio]{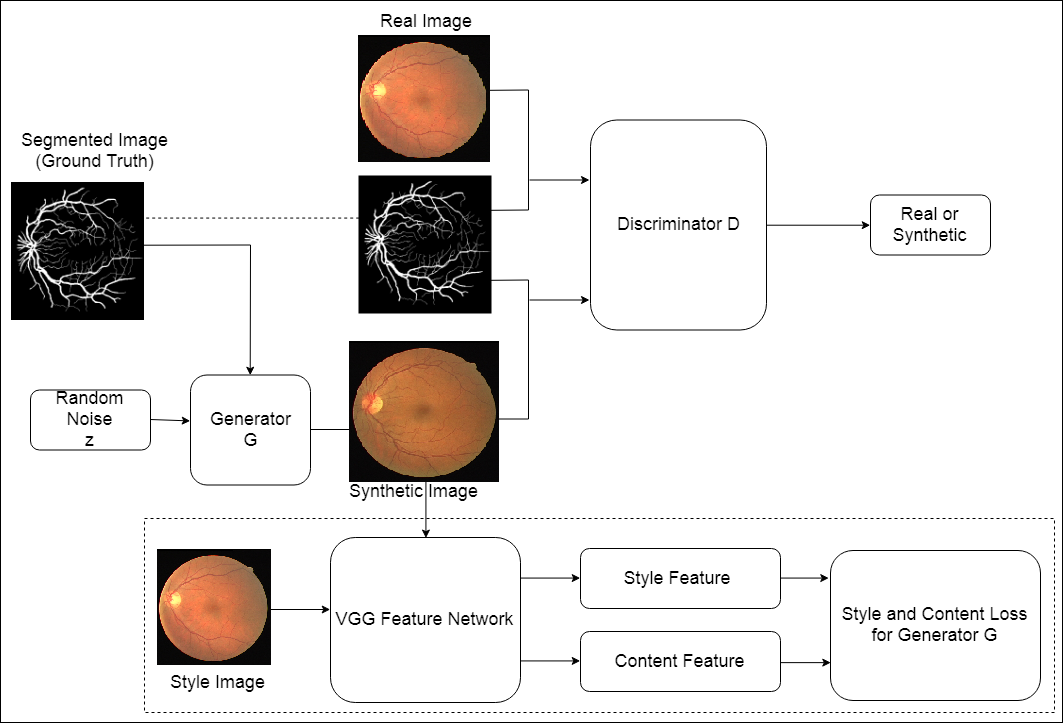}
		\caption{Flowchart of our method}
		\label{fig:a1}
\end{center}
\end{figure*}
\section{Generative Adversarial Network for Medical Imaging (MI-GAN)}
We generate segmented images using ground truth segmented images of each dataset. To produce realistic filamentary structured output, we imitate image formation process $G_{\theta}$ i.e. image generation function. Input to this function is segmented binary image $y$ and normally distributed noise $z$. Our goals are: 
\begin{enumerate}
	\item Learn $G_{\theta}$ function from very small training set.
	\item Explore conditional probability of image formation distribution $p(X|y)$. Here $X$ is random variable used to show feasible image realization conditioning for any particular realization $y$. In simple words, by varying noise vector $z$, we should get plausible as well as distinct RGB image from same segmented input $y$.
	\item Add these synthesized images to training set and improve the overall performance of the supervised segmentation.
	\item Interesting thing about our method is that a specific image style learned from an additional input $x_{s}$ is directly transferred to output image $\hat{x}$. Note that the style of the $x_{s}$ can be different from original image $x$. Similarly, their corresponding segmented images $y_{s}$ and $y$ are also unrelated. 
\end{enumerate}
The achievement of these goals is challenging as image generation process is complex process and $G_{\theta}$ is a sophisticated function. Nonetheless, using a powerful deep learning methodology i.e. GANs, an end to end machine learning algorithm is proposed in this work. Figure \ref{fig:a1} shows the overall flow of our proposed approach.\\
Along with generator $G_{\theta}$, we have discriminator $D_{\gamma}$ which gives output [0 $or$ 1] depending on the input. Discriminator function is to classify synthetic image as 0 or synthetic and real image as 1 or real. Mathematically: $X:=\hat{x}$ i.e. generated image then (d $\rightarrow$ 0) and if $X:=x$ i.e. real image from dataset then (d $\rightarrow$ 1) (see Figure \ref{fig:a1}). Here $d$ is discriminator output.\\
The training mechanism of GANs can be considered as two players competing against each other in a min-max game. Each player wants to get better than other and ultimately become the winner. Based on this analogue we define the optimization problem characterizing the G and D interplay, as:
\begin{equation}
\label{eq:a1}
\begin{split}
\stackrel{min}{\theta}\ \stackrel{max}{\gamma} \;L\;(G_{\theta}, D_{\gamma}) \;= \;E_{x,y \Rightarrow p(x,y)} \;[log D_{\gamma}(x,y)] \;\\+ \:E_{y\Rightarrow p(y),z \Rightarrow p(z)} \;[log(1 - D_{\gamma}\:(G_{\theta}(y,z),y))] \;\\+ \;\lambda L_{DEV}(G_{\theta})
\end{split}
\end{equation}
Here $\lambda$ is a trade-off constant and $\lambda$ $> 0$. This last term is introduced to make sure that the synthetic image produced by the generator is not too much deviated from the real image. It can be considered as simple L1 loss function, denoted as:
\begin{equation}
\label{eq:a2}
L_{DEV} \; (G_{\theta}) = E_{x,y \Rightarrow p(x,y)} \; [\left\|x - G_{\theta} (y,z)\right\|_{1}]
\end{equation}
During the training, generator tries to generate realistic looking synthesized images so that it may fool the discriminator and let the discriminator classify these generated images as real ones. The generator achieves this by minimizing equation \ref{eq:a1}, which is our objective function. Practically, by using the approximation scheme as in \cite{b17}, this can be done by minimizing $-log \; (D_{\gamma}  (G_{\theta}(y,z)))$, which is a simpler form than original $log \; (1- D_{\gamma} \; (G_{\theta}(y,z)))$. Overall generator loss can be defined as:
\begin{equation}
\label{eq:a3}
L_{G} (G_{\theta}) = -\sum_{i} \; log \: D_{\gamma}  (G_{\theta} (y_{i},z_{i}),y_{i}) \; + \; \lambda \left\|x_{i} \; - G_{\theta} \; (y_{i},z_{i})\right\|_{1}
\end{equation}  
On the other side, discriminator $D$ tries to properly classify and separate synthesized images from the real images by maximizing the objective function (see equation \ref{eq:a1}). The discriminator loss is determined by:
\begin{equation}
\label{eq:a4}
L_{D} (D_{\gamma}) = \sum_{i} \; log \: D_{\gamma}  (x_{i},y_{i}) \; + \; log \; (1 - D_{\gamma} \; (G_{\theta} (y_{i},z_{i}),y_{i})).
\end{equation}
The empirical summation is used to approximate the expectation value. The training is done by alternating the optimization operation between the generator and discriminator objective function. This is same as adopted by different GANs \cite{b17}, \cite{b11}, \cite{b20}.  Unfortunately, these GANs do not provide a formal guarantee that this optimization process will converge and we will be able to reach at Nash Equilibrium point. Different tricks are available which guarantee convergence of GANs training process and produces reasonable realistic looking synthesized images at output \cite{b17}, \cite{b11}, \cite{b20}. Figure \ref{fig:a1} illustrates overview of the work flow of proposed GAN model, excluding the dashed box. Next we discuss specific neural network architecture of our Generator G and Discriminator D in detail.
\subsection{Generator and Discriminator Architecture in MI-GAN}
\begin{figure*}[]
 \begin{center}
	\includegraphics[width=1.3\textwidth, height= 10cm, angle=90]{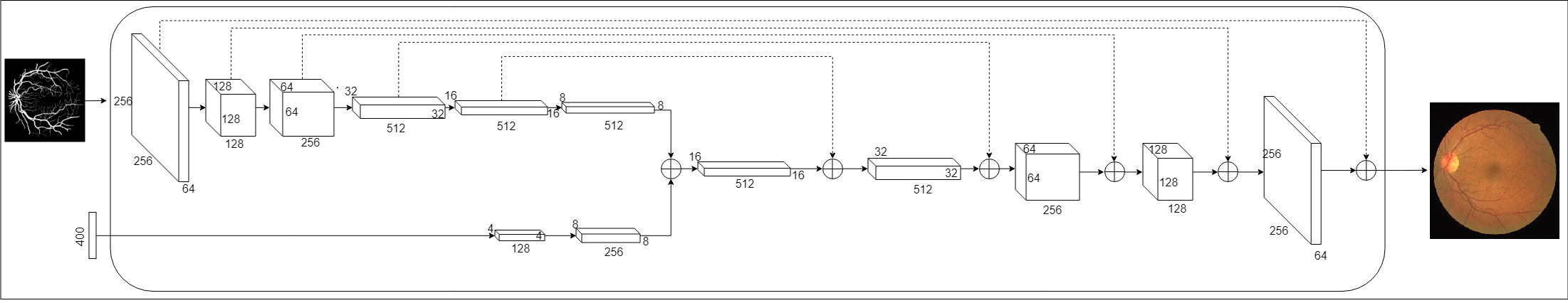}
	 \caption{Generator Structure}
	\label{fig:a2}
\end{center}
\end{figure*}

\begin{figure}[]
  \includegraphics[width=8cm, height=5cm]{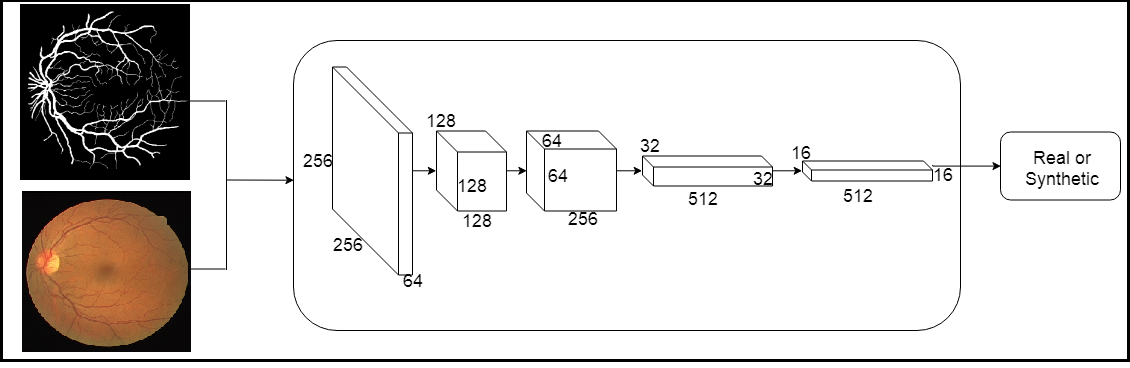}
  \caption{Discriminator Structure}
	\label{fig:a3}
\end{figure}

Explained in \cite{b15}, \cite{b22},\cite{b23} and \cite{b24}, commonly used technique of encoder-decoder is adopted here. This allows us to introduce noise code in natural manner. Encoder acts as feature extractor. It is a multiple layered neural network which captures local data representation in first few layers and goes on to capture more global representation as we move deep inside the neural network. A 400 dimensional random noise code z is fully connected to first layer of the network (see Figure \ref{fig:a2}). This noise code is then reshaped. One thing to note is that for all the layers of G and D, we use kernel with fixed size and there are two strides with no pooling layers. Meanwhile in our case, it is important for the generator to respect morphology of input segmented image while generating output images. To do so, the `skip connections' of U-Net \cite{b16} are taken into consideration. In skip connections approach the previous layer is mirrored and then duplicated by appending it to the current layer. Odd numbered layers are skipped and the center coding is considered as origin. Note that if we have small image size and a deep neural network, the encoder-decoder framework does work well even without using skip connections. However, we are working with $512 \times 512$ sized images (which is a large size) and our network is relatively deep.\\
Training such a model is challenging. The main challenge one may face while using deeper network is of vanishing gradient over a long path during error back-propagation. `Skip connections', used similarly as in residual nets \cite{b25}, allows us to pass the error gradients directly from decoder layer to its corresponding encoder layer. This facilitates the memorization of local and global shapes representation as well as their corresponding textures encountered in training dataset, thus we are able to generate better results. We use the basic architecture of the network proposed in \cite{b11} to build layers of our generator having multiple convolution layers, Batch Normalization and Leaky ReLU components as shown in Figure \ref{fig:a2}. The activation function used to squash the output of the final layer is $tanh$. This function limits the output value between 1 and -1. 
With our generator, the discriminator network is also built by convolution layers, Batch Normalization and Leaky ReLU, as shown in Figure \ref{fig:a3}. The activation function used at output layer is $sigmoid$ instead of $tanh$. After every convolution the feature map size is halved. For example, as we have input image of $512 \times 512$ so after one convolution layer image size will be decreased to $256 \times 256$. The number of feature maps (filters) are doubled from 32 through 512 as we move from first to last layer.\\
Uptill here, we have explained how our proposed approach learns the generic representation from a small training dataset and use it to employ generation of synthesized segmented images. Next, we discuss the segmentation process and a variant of style transfer technique.
\subsection{Segmentation Technique}
For segmentation, we utilize gold standard segmented images. We add a loss function that penalizes the distance between gold standard images and output segmented images. This loss is defined as binary cross-entropy i.e.
\begin{equation}
\label{eq:a5}
L_{SEG}\:(G_{\theta}) = E_{x,y \Rightarrow p(x,y)}\: - \: y \; log G_{\theta}(x)\: -\: (1-y) \; log(1-G_{\theta}(x)).
\end{equation}
The objective function can be formulated by summing the GAN objective function and segmentation loss. So, the new objective function is as follows:
\begin{equation}
\label{eq:a6}
L_{G}(G_{\theta})\: = -\sum_{i} log D_{\gamma}\:(G_{\theta}(y_{i},z)) + \lambda L_{SEG}(G_{\theta}).
\end{equation}  
$\lambda$ is used to balance both the objective functions.
\subsection{Style Transfer Variant for MI-GAN}
Recent advancement in image style transfer, such as \cite{b21}, inspired us to use this technique in the field of medical imaging. Here given an input segmentation image $y$ which delineates content of its filamentary structure, we expect that the generated image $\hat{x}$ possess the unique texture (referred as style) of the input $x_{s}$ which is our target, while still adhering the content of $y$ presented during the training. The difference of our style transfer approach from original style transfer is that instead of generic representation, our synthesized image is based on a particular style representation provided by $x_{s}$. The procedure we follow is that we introduce a style image as an additional input along with training input i.e. a new segmented image $x_{s}$ is introduced having different style and texture. Note that in general $x_{s}$ has its own filamentary structure (segmentation), which is different from other input $y$. Nonetheless, this does not affect the performance of generating synthesized images using our method. It is worth noticing that the proposed methodology is practically implementable in biomedical imaging field. On one hand we have very less annotated images available while on the other hand there are a lots of unannotated images available on world wide web  which can be used as potential style inputs.\\
The overall training and testing methodology of this new algorithm is same as we have described in our approach. The training is carried out in form of batches for all $n$ annotated examples in training set. The generator and discriminator is same as mentioned before but the only difference is that in objective function (see equation \ref{eq:a1}), a new cost term $L_{ST}(G_{\theta})$ is introduced, which replaces $\lambda L_{DEV}(G_{\theta})$ in equation \ref{eq:a1}. We follow style transfer idea proposed in \cite{b7}, \cite{b8} to use the Convolutional Neural Network (CNN) of VGG-19 \cite{b26} for extraction of the features from this multi-layered network. VGG-19 network architecture is basically a series of five CNN blocks of VGG net. Each block further consists of two to four consecutive CNN layers of same size. Let us define some notations for convenience. Let $\Gamma$ be the index for a set of CNN blocks and $\gamma$ is the index of a particular block where $\gamma \in \Gamma $. Set of layers be represented by $\Lambda(\gamma)$ or $\Lambda$. Here layer index is $\lambda$ such that $\lambda \in \Lambda$. Now the segmented image $X$ is denoted as $\phi_{\gamma}^{\lambda}(X)$, irrespective of real image $x$ or generated $\hat{x}$. VGG-19 network is obtained by training the \textit{ImageNet omega classification} problem, which is explained in detail in \cite{b26}. Optimization problem for this style transfer algorithm is explicitly incorporated with two perceptual losses i.e. style loss and content loss of \cite{b6}, as well as total variational loss.\\
\textbf{Style loss:} This loss is used to minimize total textural deviation between target style $x_{s}$ and generated image $\hat{x}$. To calculate this loss, consider $\Gamma_{s}$ showing set of CNN blocks and for each block $\gamma_{s} \in \Gamma_{s}$. The set of layer is represented by $\Lambda_{s}$. $\lambda_{s}$-th layer of $\gamma_{s}$ block is defined as $\phi _{{\gamma _s}}^ {{\lambda _s}}\left( X \right)$. Here, $X = \hat{x}$ or $X = x_{s}$.  Total number of interest feature maps inside current layer $\lambda_{s}$ is denoted by $\left| \lambda_{s} \right|$. Let $i$ and $j$ be index of interest feature map and $k$ be index of an element of current feature map. Information of the corresponding feature is characterized using Gram matrix $G^{\lambda_{s}}_{\gamma_{s}}(X)$ which belongs to $R^{\left| \lambda_{s} \right| \times \left| \lambda_{s} \right|}$. Each element $G^{\lambda_{s}}_{\gamma_{s},ij}(X)$ is defining an inner product of $i^{th}$ and $j^{th}$ interest feature maps in $\lambda_{s} ^ {th}$ layer of block $\gamma_{s}$. Mathematically, 
\begin{equation}
\label{eq:a6}
G^{\lambda_{s}}_{\gamma_{s},ij} = \sum_{k} \phi^{\lambda_{s}}_{\gamma_{s},ik} \ \: \phi^{\lambda_{s}}_{\gamma_{s},jk}.
\end{equation}
The style loss of $x_{s}$ and $\hat{x}$ during training is defined as:
\begin{equation}
\label{eq:a7}
\begin{split}
l_{sty} \; (G_{\theta}) = \sum_{\gamma_{s} \in \Gamma_{s}, \lambda_{s} \in \Lambda_{s}} \: \frac{\varpi_{\gamma_{s}}}{W_{\gamma_{s}} \: H_{\gamma_{s}}} \; \times \\ {\left\|G^{\lambda_{s}}_{\gamma_{s}}(x_{s}) - G^{\lambda_{s}}_{\gamma_{s}}(\hat{x}) \right\|^2}_{F} . 
\end{split}
\end{equation}
Here $\left\|.\right\|_{F}$ is matrix Frobenius norm, $\varpi_{\gamma_{s}}$ represents weight of $\gamma_{s}$-th block Gram matrix. Note that by definition $\hat{x}$ = $G_{\theta}(y,z)$.\\
\textbf{Content loss:} Following notations are considered for content loss: $\Gamma_{c}$ is index of set of convolution neural network blocks while each block index is as $\gamma_{c} \in \Gamma_{c}$. Set of layers is represented as $\Lambda_{c}$. We expect the synthesized output $\hat{x}$ will abide the segmentation pattern of the real image (input image) $x$. To make sure this happens, we encourage output image to minimize the Frobenius norm of the difference between input and output CNN features. Mathematically,
\begin{equation}
\label{eq:a8}
l_{cont}(G_{\theta}) \: = \: \sum_{\gamma_{c}\in\Gamma_{c},\lambda_{c}\in\Lambda_{c}} \frac{1}{W_{\gamma_{c}}H_{\gamma_{c}}}\:{\left\|{\phi_{\gamma_{c}}^{\lambda_c}} (x) - {\phi_{\gamma_{c}}^{\lambda_c}} \hat{(x)}\right\|^2}_{F}.
\end{equation}
\textbf{Total variational loss:} Total variational loss is incorporated using following equation for spatial smoothness of the generated images.
\begin{equation}
\label{eq:a9}
l_{tv}(G_{\theta})\: = \sum_{w,h} ({{\left\| \hat{x}_{w,h+1} - \hat{x}_{w,h}\right\|}^2}_{2} \: + \: {{\left\| \hat{x}_{w+1,h} - \hat{x}_{w,h}\right\|}^2}_{2}).
\end{equation}
Here $\hat{x}_{w,h}$ denotes pixel value of location in generated image $\hat{x}$, where ${w,h} \in {W,H}$ respectively. Summarizing all the three loss functions combined together gives us \textit{Style Loss} $L_{ST} (G_{\theta})$,
\begin{equation}
\label{eq:a10}
L_{ST}(G_{\theta}) = \omega_{cont}l_{cont}\:+\:
\omega_{sty}l_{sty}\:+\:\omega_{tv}l_{tv}.
\end{equation}
So, now we modify $L_{DEV}$ in equation 1 by this style transfer loss $L_{ST}$. The new objective function for generator G becomes:
\begin{equation}
\label{eq:a11}
L_{G}(G_{\theta})\: = -\sum_{i} log D_{\gamma}\:(G_{\theta}(y_{i},z)) + L_{SEG}(G_{\theta}) + L_{ST}(G_{\theta}).
\end{equation}
Discriminator objective function remains unchanged (see equation \ref{eq:a4}). Style transfer from input style $x_{s}$ is obtained using back-propagation optimization of the above objective function.
\section{Experimental Setup}
\subsection{Datasets Preparation}
\begin{figure}[]
  \begin{center}
	\includegraphics[width=9cm,height=10cm, keepaspectratio, angle = -90]{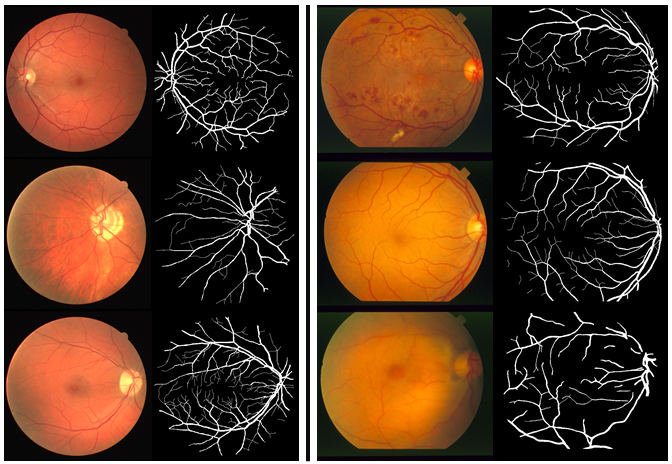}
	\caption{(\textbf{From Top to Bottom}) DRIVE Dataset Images with their ground truth and STARE Dataset Images with their ground truth.}
	\label{fig:a4}
\end{center}
\end{figure}

For evaluation of the proposed approach, we use two benchmark datasets. The first is DRIVE dataset and the second is STARE dataset. These both datasets include a broad spectrum of vascular structured retinal images. The image sizes and number of training examples are different in each dataset. In DRIVE dataset there are 20 training examples with image size of $584 \times 565$ while STARE dataset has 10 training images with image size of $700 \times 605$. The images in both the datasets are roughly similar. In pre-processing stage all the images are re-sized to $512 \times 512$. Images in DRIVE dataset contain large size background area thus they are cropped into $565 \times 565$ sized sub-image centered to the original one to make sure all the fore-ground pixels are still contained in the new image. Then this image is again re-sized to $512 \times 512$ using bi-cubic interpolation. Images in STARE dataset has rather small background margins (area outside fore-ground mask) so they are directly converted to $512 \times 512$ using bi-cubic interpolation. Pixel values of all the input signals are scaled down in-between -1 and 1 so that our generator should learn to generate synthetic image of size $512 \times 512$. In the last stage these images are again up-sampled to there original sizes. The final result is obtained by applying circular mask to the segmented image so that only inside pixels are retained as a fore-ground. Figure \ref{fig:a4} shows few fundoscopic images from DRIVE (upper row) and STARE (lower row) along with their ground truths.
\subsection{Parameters of proposed model}
The 3D boxes in generator as well as in discriminator (in Figure \ref{fig:a2} and \ref{fig:a3}) shows CNN layer with its number of features size. Edges of the boxes show the convolutional or de-convolutional operation having filter size of $w_{f} \times h_{f} \times l_{f}$. Here we have considered $4 \times 4 \times l_{f}$, where $l_{f}$ is self-manifested by third dimension of consecutive layer. The number in the figure \ref{fig:a2} and \ref{fig:a3} specify intrinsic parameters of the networks. For example, length of the noise vector is 400 and size of first layer is $256 \times 256 \times 64$. In generator G, the sign $\oplus$ along with two directed edges pointing inward shows concatenation operation. Let us see the first $\oplus$; here concatenation operation takes place between $8 \times 8 \times 256$ tensor and $8 \times 8 \times 512$ tensor to produce $8 \times 8 \times 786$ tensor. This concatenation operation is followed by deconvolution operation using filter size of $4 \times 4 \times 512$ which in-return produces 3-D box with size of $16 \times 16 \times 512$.\\
We update generator G twice and then update discriminator D during the learning iteration to balance the overall learning process of generator and discriminator. Noise is sampled element-wise from zero mean Gaussian having standard deviation of 0.001 during training. Standard deviation is changed to 1 and sampling is done in same manner as above,  when we evaluate our algorithm. Based on observation, this change in standard deviation is useful to maintain proper level of diversity as we have very small-size data. To get better training of generator and discriminator in our model, batch normalization \cite{b27} is used right after every convolutional layer.\\
The VGG-19 network is used to produce feature descriptor for style transfer algorithms. Output of this network is style and content features. Values of the parameters used in this network are: $\Gamma_{s} = {1,2,3,4,5}$ and $\Gamma_{c} = 4$. $\Lambda_{s} = 1$ for style loss and $\Lambda_{c} = 0$ for content loss. $\varpi_{\gamma_{s}}$ is kept fixed for all blocks and is set to 0.2. The weights of three loss functions are as follow: $\omega_{cont} = 1$, $\omega_{sty} = 10$ and $\omega_{tv} = 100$.\\
Images are augmented by rotation and left-right flip and then normalization is done on each image to get z-score for each channel. These augmented images are then divided in train and validation images with ratio of 19 to 1. Generator having least validation loss is selected from the models. The generator and discriminator are trained for n epochs until convergence. For optimization of the objective function we use Adam optimizer. The learning rate is set to $2e^{-4}$ and trade-off co-efficient $\lambda = 10$.\\
All the experimentation is carried out using standard PC with Intel Core i5 CPU and GeForce GTX 1080 GPU with 8 GB memory. We evaluate our technique with Area Under Curve for Precision and Recall (AUC PR), Dice co-efficient (F1-score) and Area Under Curve for Receive Operating Characteristics (AUC ROC). The probability map is threshold using Ostu thresholding \cite{b28}, which is mostly used to separate fore-ground from background for calculation of dice co-efficient. Pixels inside the Field Of View (FOV) is counted when we are computing the measures, for fair measurement.  
\begin{table}[h]
\centering
\caption{Comparison of different models having different discriminators}
\label{Tab: comp}
\begin{tabular}{|l|l|l|l|l}
\hline
\multirow{2}{*}{Model} & \multicolumn{2}{l|}{DRIVE}      & \multicolumn{2}{l|}{STARE}                           \\ \cline{2-5} 
                       & ROC            & PR             & ROC            & \multicolumn{1}{l|}{PR}             \\ \hline
U-Net \cite{b26}                 & 0.970          & 0.886          & 0.973          & \multicolumn{1}{l|}{0.902}          \\ \hline
Pixel GAN   \cite{b36}           & 0.971          & 0.889          & 0.967          & \multicolumn{1}{l|}{0.897}          \\ \hline
Patch GAN-1 (10x10) \cite{b15}    & 0.970          & 0.889          & 0.976          & \multicolumn{1}{l|}{0.903}          \\ \hline
Patch GAN-2 (80x80) \cite{b15}  & 0.972          & 0.893          & 0.977          & \multicolumn{1}{l|}{0.908}          \\ \hline
Image GAN \cite{b35}             & \textbf{0.980} & \textbf{0.914} & \textbf{0.983} & \multicolumn{1}{l|}{\textbf{0.916}} \\ \hline
\end{tabular}
\end{table}
\begin{table*}[]
\centering
\caption{Comparison of proposed method with other existing techniques on basis of AUC ROC and PR and Dice Score}
\label{tab: results}
\begin{tabular}{|c|c|c|c|c|c|c|}
\hline
\multirow{2}{*}{Method} & \multicolumn{3}{c|}{DRIVE}                       & \multicolumn{3}{c|}{STARE}                       \\ \cline{2-7} 
                        & Dice Score     & AUC ROC        & AUC PR         & Dice Score     & AUC ROC        & AUC PR         \\ \hline
Our Method              & \textbf{0.832} & \textbf{0.984} & \textbf{0.916} & \textbf{0.838} & \textbf{0.985} & \textbf{0.922} \\ \hline
Kernel Boost  \cite{b1}  & 0.800          & 0.931          & 0.846          & -              & -              & -              \\ \hline
$N^{4}$ - Fields \cite{b3}    & 0.805          & 0.968          & 0.885          & -              & -              & -              \\ \hline
DRIU   \cite{b31}         & 0.822          & 0.979          & 0.906          & 0.831          & 0.972          & 0.910          \\ \hline
Wavelets \cite{b32}       & 0.762          & 0.943          & 0.814          & 0.774          & 0.969          & 0.843          \\ \hline
HED     \cite{b33}        & 0.796          & 0.969          & 0.877          & 0.805          & 0.976          & 0.888          \\ \hline
Human Expert          & 0.791          & -              & -              & 0.76           & -              & -              \\ \hline
\end{tabular}
\end{table*}
\section{Results and Discussions}
\begin{figure}[]
\begin{center}
  \includegraphics[width=\textwidth,height=8cm, keepaspectratio]{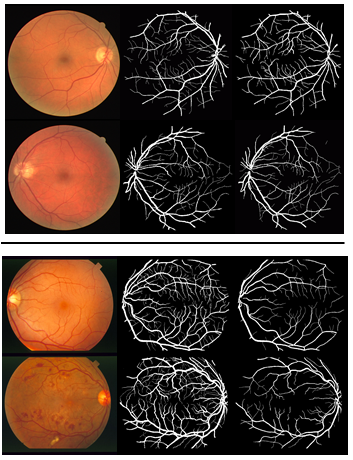}
  \caption{Fundoscopic images (first column), Probability Map of DRIU (second column) and Probability Map of Our Method (third column). Top image is DRIVE dataset and Bottom image is STARE dataset.}
\label{fig:a5}
\end{center}
\end{figure}

\begin{figure}[]
\begin{center}
  \includegraphics[width=\textwidth,height=8cm, keepaspectratio]{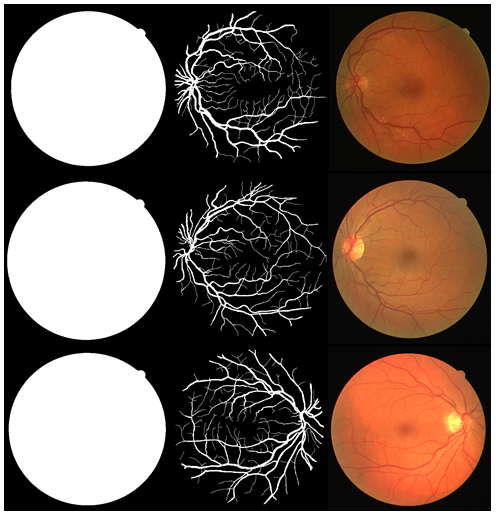}
  \caption{(\textbf{From left to right}) Masks, filamentary structures and Output retinal images}
	\label{fig:a6}
\end{center}
\end{figure}

In Table 1, we have compared performance of different models with different discriminators. There is no discriminator in U-Net so it shows inferior performance as compare to patch GAN and Image GAN. Patch GAN and Image GAN have shown improvement in overall segmentation quality but Image GAN, which has most powerful discriminator framework, out-performs all the other networks. This result is enough to claim that a powerful discriminatory framework is key for successful training of the networks with GANs \cite{b29},\cite{b30}.\\
Table 2 summarizes dice coefficients (F1-score), AUC for ROC and AUC for PR for our proposed method in comparison with other methods. Our method outperformed all the existing methods and shows better dice coefficient and AUC values. Our method also surpasses human's annotating ability on DRIVE dataset.\\
Qualitative comparison of segmentation using our method and best existing method DRIU (Deep Retinal Image Understanding \cite{b31}) is illustrated in Figure \ref{fig:a5}. Our proposed method generates concordant probability values to the gold standard while DRIU gives overconfident probability on boundaries between vessels and background, as well as on fine vessels. This may cause over-segmentation of retinal image, resulting in high false positive values. In contrast, the proposed technique allows more false negatives near the edges and terminal end of the vessels because it has tendency to give low probability to the pixels which falls in uncertain region. This is same as human annotators would do.
In Figure \ref{fig:a6}, we have shown the generated masks (outer boundary), filamentary structured image and generated output images. We can see that these generated output images are visually close to real ones.\\
\section{Conclusion}
In this paper, we have introduced a new Generative Adversarial Network for Medical Imaging (MI-GAN) framework which focuses on retinal vessels image segmentation and generation. These synthesized images are  realistic looking. When used as additional training dataset, the framework helps to enhance the image segmentation performance. The proposed model is capable of learning useful features from a small training set. In our case the training set consisted of only 10 examples from each dataset namely DRIVE and STARE. Our model outperformed other existing models in terms of AUC ROC, AUC PR and Dice co-efficient. Our method had less false positive rate at fine vessels and have drawn more clearer lines, as compared to other methods. Future work involves investigation into datasets of different bio-medical images for interplay of synthesized images, domain adaptation tasks and segmentation of the medical images. 
\section*{Compliance with Ethical Standards}
Funding: No funding declared.\\
Conflict of Interest: Talha Iqbal declares that he has no conflict of interest. Hazrat Ali declares that he has no conflict of interest.\\
Ethical approval: This article does not contain any studies with human participants or animals performed by any of the authors.\\
Informed consent: Not applicable.
\bibliographystyle{elsarticle-num}
\bibliography{ref}
\end{document}